\documentclass[10pt,twocolumn,letterpaper]{article}

\usepackage[pagenumbers]{cvpr} 

\definecolor{cvprblue}{rgb}{0.21,0.49,0.74}
\usepackage[pagebackref,breaklinks,colorlinks,allcolors=cvprblue]{hyperref}
\usepackage{algorithm}
\usepackage{algorithmic}
\usepackage{booktabs}
\usepackage{algorithm}
\usepackage{algorithmic}
\usepackage{amsmath}
\usepackage{amssymb}
\usepackage{subcaption}
\usepackage{bbding}
\usepackage{diagbox} 
\usepackage{array}  
\usepackage{multirow}
\usepackage[table]{xcolor}
\def\paperID{****} 
\def\confName{CVPR}
\def\confYear{2026}

\title{Visual Document Understanding and Reasoning: A Multi-Agent Collaboration Framework with Agent-Wise Adaptive Test-Time Scaling}

\author{
   Xinlei Yu$^{1}$ $\quad{}$ Chengming Xu$^{2}$ $\quad{}$ Zhangquan Chen$^{3}$ $\quad{}$ Yudong Zhang$^{3}$ $\quad{}$ Shilin Lu$^{5}$ \\
   $\quad{}$ Cheng Yang$^{7}$ $\quad{}$ Jiangning Zhang$^{6}$ $\quad{}$ Shuicheng Yan$^{1}$ $\quad{}$ Xiaobin Hu$^{1}$\thanks{Corresponding authors.} \\
\small$^1$National University of Singapore \quad 
  $^2$Tencent Youtu Lab \quad 
  $^3$Tsinghua University  \\
  \small$^4$University of Science and Technology of China \quad
  $^5$Nanyang Technological University \quad
  $^6$Zhejiang University \quad  
  $^7$DeepWisdom\\ 
}

\begin{document}
\maketitle


\begin{abstract}
The dominant paradigm of monolithic scaling in Vision-Language Models (VLMs) is failing for understanding and reasoning in documents, yielding diminishing returns as it struggles with the inherent need of this domain for document-based procedural reasoning, cognitive complexity, and factual accuracy.
To this end, we introduce \textbf{MACT}, a \textbf{M}ulti-\textbf{A}gent \textbf{C}ollaboration framework with agent-wise adaptive \textbf{T}est-time scaling that pioneers a paradigm shift to procedural scaling, adapting dynamically to the functional entities of visual documents understanding and reasoning. MACT decomposes the visual document processing flow into four specialized agents, \textit{i.e.}, planning, execution, judgment, and answer, to resolve cognitive overload and introduce a critical self-correction loop for factual grounding.
This collaborative architecture is amplified by an agent-wise adaptive test-time scaling strategy that intelligently allocates computational resources based on the complexity and redundancy of each functionality. 
Evaluated on multiple visual document understanding benchmarks, MACT achieves superior performance with a smaller parameter scale, adapting effectively to various document scenarios without compromising its general or mathematical reasoning capabilities. The three variants of MACT consistently attain top-three average performance rankings, with average performance enhancements of 9.9–11.5\% over the base models. The source code will be released publicly.

\end{abstract}    

\vspace{-5pt}
\section{Introduction}

For humans, acquiring and understanding visual information from documents is an indispensable part of daily life. The corresponding tasks of visual documents, \textit{e.g.}, Visual Question Answering (VQA), serves as a quantifiable benchmark for a model's ability to comprehend complex layouts, dense text, and structured data. To this end, the field has seen a rapid evolution of benchmarks, moving from simple documents to multi-page reports, complex charts, and webpages. The questions have similarly evolved from simple lookups to requiring semantic understanding and multi-hop reasoning, pushing the capability boundaries of VLMs.

\begin{figure*}[t]
    \centering 
    \begin{subfigure}[b]{0.315\textwidth} 
        \centering
        \includegraphics[width=\textwidth]{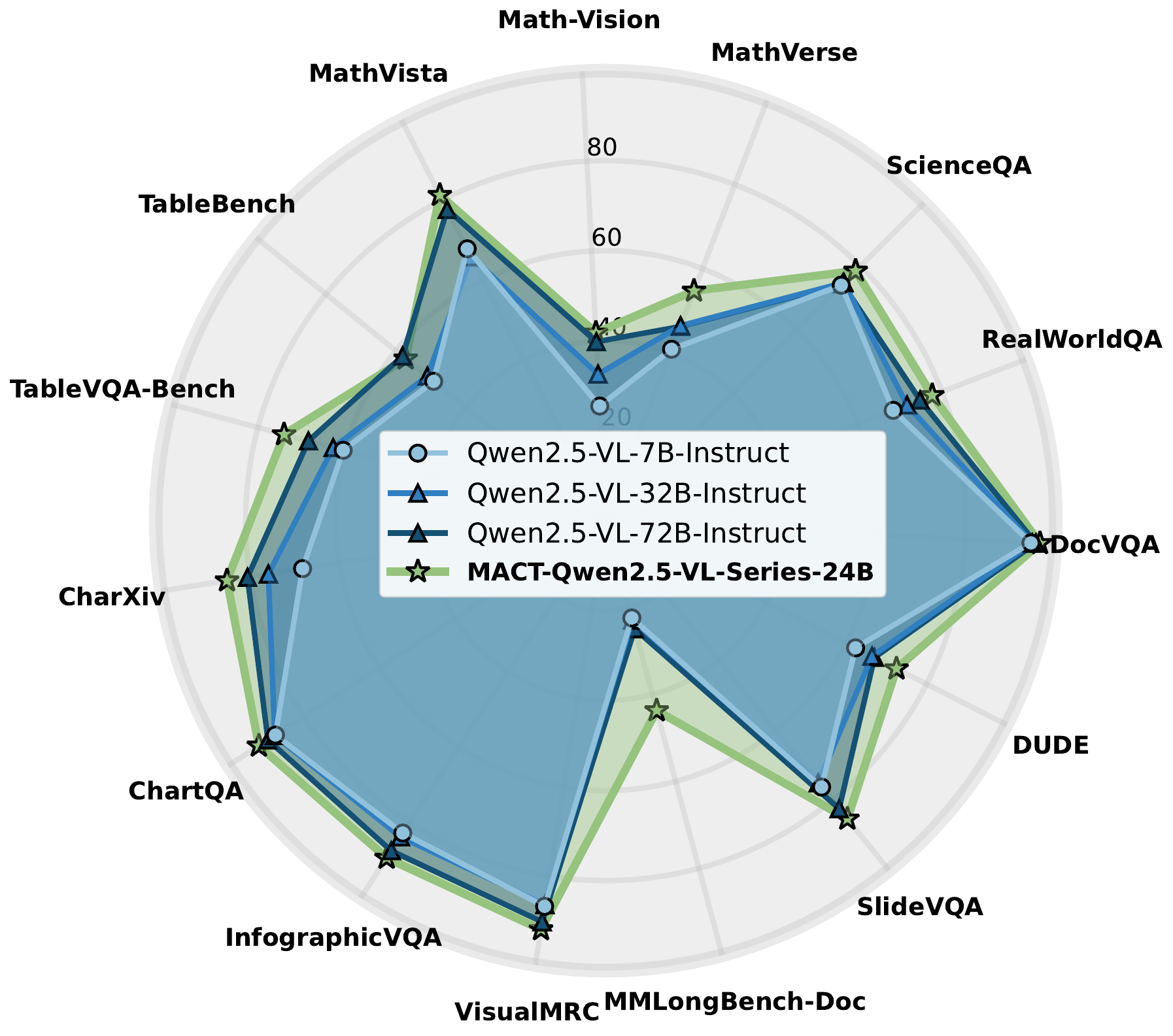}
        \subcaption{Comparisons on Qwen2.5-VL Series~\cite{qwen2025qwen25technicalreport}} 
    \end{subfigure} 
    \hfill 
    \begin{subfigure}[b]{0.315\textwidth}
        \centering
        \includegraphics[width=\textwidth]{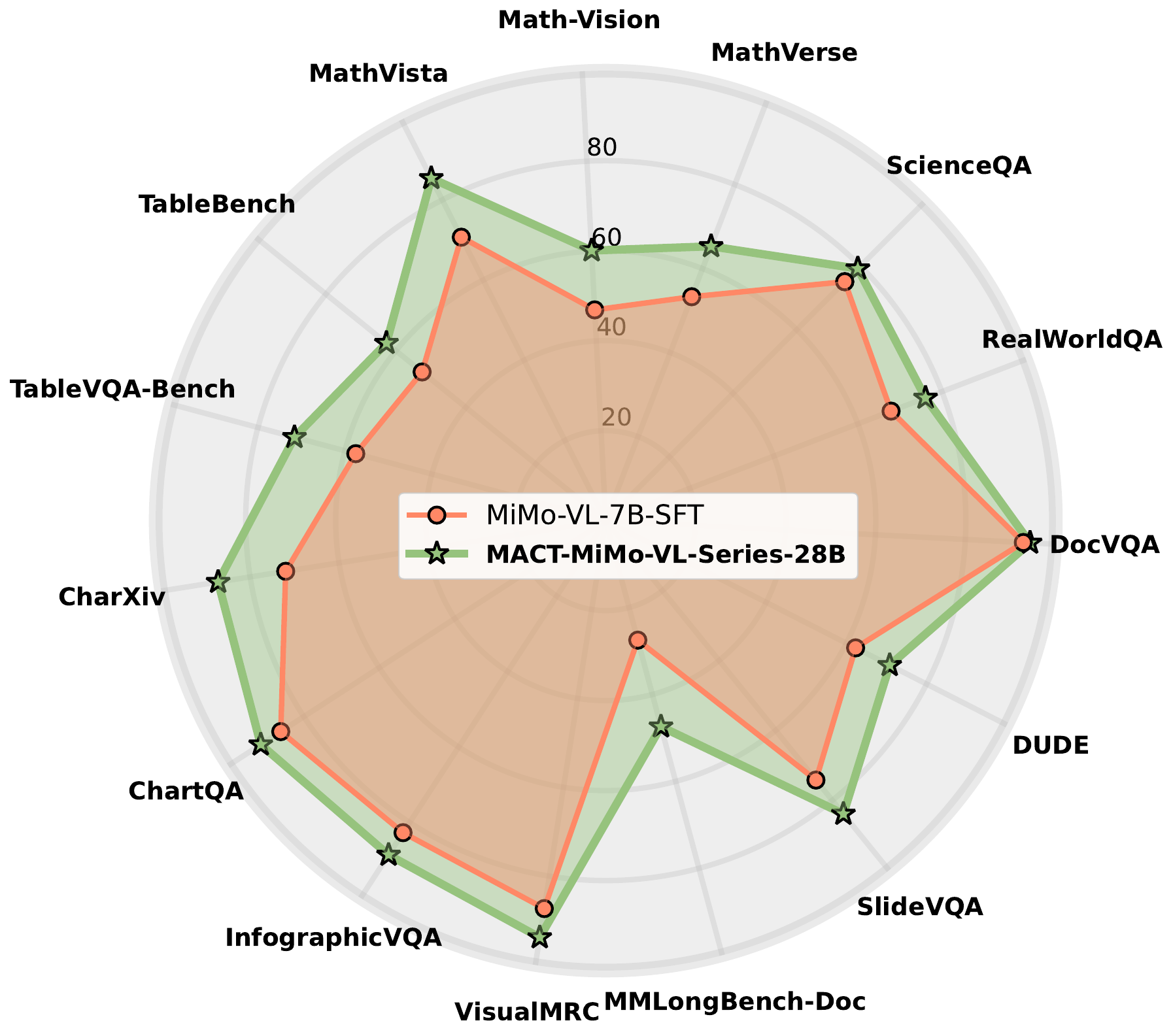}
        \subcaption{Comparisons on MiMo-VL Series~\cite{xia2025mimo}}
    \end{subfigure}
    \hfill 
    \begin{subfigure}[b]{0.315\textwidth}
        \centering
        \includegraphics[width=\textwidth]{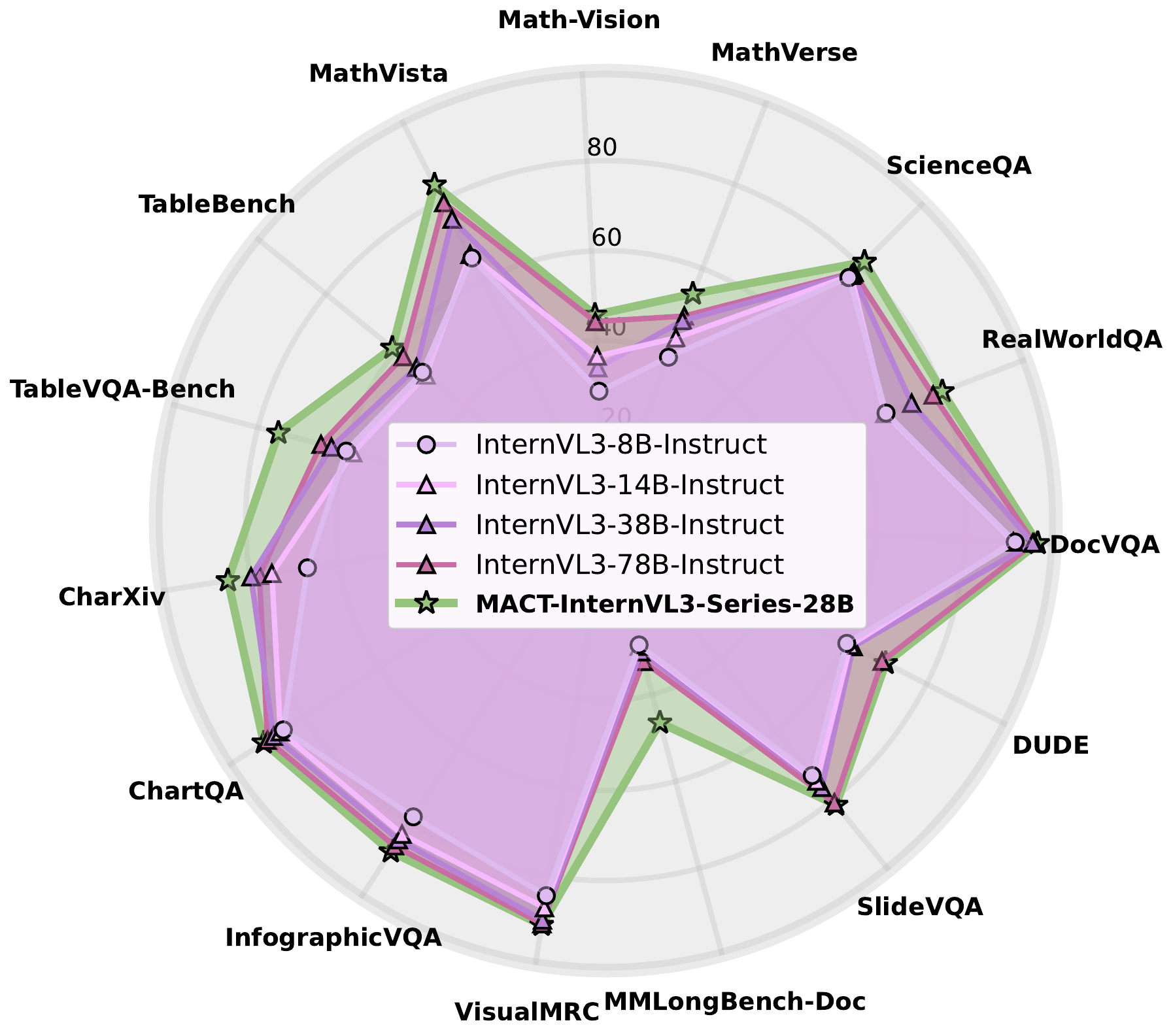}
        \subcaption{Comparisons on InternVL-3 Series~\cite{zhu2025internvl3}}
    \end{subfigure}

    \caption{Comparisons among the three variants of MACT, the base models of these variants, and larger-scale models within the same model family, indicating the superiority of our framework over monolithic scaling paradigms. } 
    \label{comparison_to_base}   
    \vspace{-10pt}
\end{figure*}

The dominant paradigm for advancing VLM capabilities, both for documents and other general vision data, has been \textbf{monolithic scaling}, \textit{i.e.}, increasing parameter size, leveraging more high-quality training data. The approach has led to remarkable success, with phenomenal models like GPT-4o~\cite{hurst2024gpt}, Gemini-2.0-Pro~\cite{gemini2024deepmind} and Claude-3.7-Sonnet~\cite{anthropic2024claude}, demonstrating powerful general-purpose abilities. However, for document-based VQA, this strategy of uniform scaling is showing diminishing returns. As shown in Fig.~\ref{comparison_to_base}, only limited performance gain is attained along parameter scaling for open-sourced models~\cite{qwen2025qwen25technicalreport,zhu2025internvl3} with exponential growth in computational overhead, suggesting that simply increasing parameter count is a brute-force solution that fails to address the core challenges of the domain.

The core problem is that \textbf{documents are not monolithic entities that are suitable for uniform scaling}. Unlike natural images, they possess a unique set of characteristics that make naive scaling less effective: \textbf{(1) Suboptimal procedural reasoning. } Document analysis is not a single action of perception but rather an inherently procedural, multi-step workflow~\cite{kutsyuruba2023document}: decomposing the question, forming a strategy, locating relevant information, and synthesizing a final answer. Monolithic models, with their single forward-pass architecture, attempt to solve this entire procedure at once, leading to less robust reasoning paths. \textbf{(2) Cognitive overload.} The procedural steps of document analysis demand a diverse set of specialized skills, such as global layout parsing, fine-grained text extraction, logical inference, and numerical calculation. A monolithic model is forced to encode all these disparate functions into a single set of weights, leading to cognitive overload and task interference. This leads to a compromise where the model achieves broad generalization at the expense of deep expertise in any single task. \textbf{(3) Vulnerability to factual errors.} The semantic meaning in documents is extremely sensitive to minor procedural mistakes~\cite{fitzgerald2012documents}, even paragraph truncation errors or misaligning a table row can invalidate the entire answer. Monolithic, feed-forward models lack an internal verification or self-correction loop. An initial extraction error, however small, inevitably cascades through the reasoning process without being challenged, identified, and corrected.

These challenges reveal the need for a new paradigm: a shift from \textbf{monolithic scaling} to \textbf{procedural scaling}, where the problem is decomposed and specialize test-time scalings are applied to each entity. To achieve this, we propose \textbf{MACT}, a \textbf{M}ulti-\textbf{A}gent \textbf{C}ollaboration framework with agent-wise adaptive \textbf{T}est-time scaling, enabling a more effective scaling framework. Specifically, it deconstructs the monolithic model into four specialized agents that explicitly mirror the required cognitive functions. Within this framework, we employ four relatively collaborative agents: planning, execution, judgment, and answer agents. This design inherently enforces procedural reasoning, dedicating a specialist to each cognitive step from planning to synthesis. This decomposition is what resolves the cognitive overload problem, allowing each agent to achieve mastery over its specific function. The key to unlocking this specialized performance is our agent-wise adaptive test-time scaling approach. Rather than uniform scaling, it is intelligent, on-demand allocation of computational resources, allowing the execution agent to apply maximum scrutiny to a single data point while the planning agent operates efficiently on the whole document. Finally, the framework directly confronts the high cost of errors. The judgment agent acts as a built-in verification, providing the self-correction capabilities that monolithic models lack.

We train three variants of MACT based on different groups of base models using a two-stage pipeline, and evaluate them on 15 benchmarks, comprising 10 document-based and 5 non-document-based ones. Our MACT showcases superior document understanding and reasoning without compromising general capabilities. Compared to models of comparable scale and larger-scale ones, the three variants of MACT demonstrate an average increase of 3.2\% to 5.6\%, securing the top three positions and achieving the best performance in 13 out of 15 benchmarks. The results demonstrate that monolithic scaling is not the optimal approach for visual document-based tasks. In contrast, the adaptive test-time scaling in multi-agent framework emerges as a more advantageous alternative, as it achieves superior scaling performance.

\section{Methodology}
This section provides a detailed elaboration of our proposed MACT, divided into three subsections. We first introduce the multi-agent framework, which decomposes the document-tailored workflow into four functionalities: planning, execution, judgment, and answer, and defines the collaboration mechanisms among them. Then, we detail the agent-wise adaptive test-time scaling and mixed reward modeling tailored to the multi-agent collaboration framework. More details are presented in Appendix~{\color{cvprblue}{7}}.


 \begin{figure*}[t]
    \centering
    \includegraphics[width=0.88\linewidth]{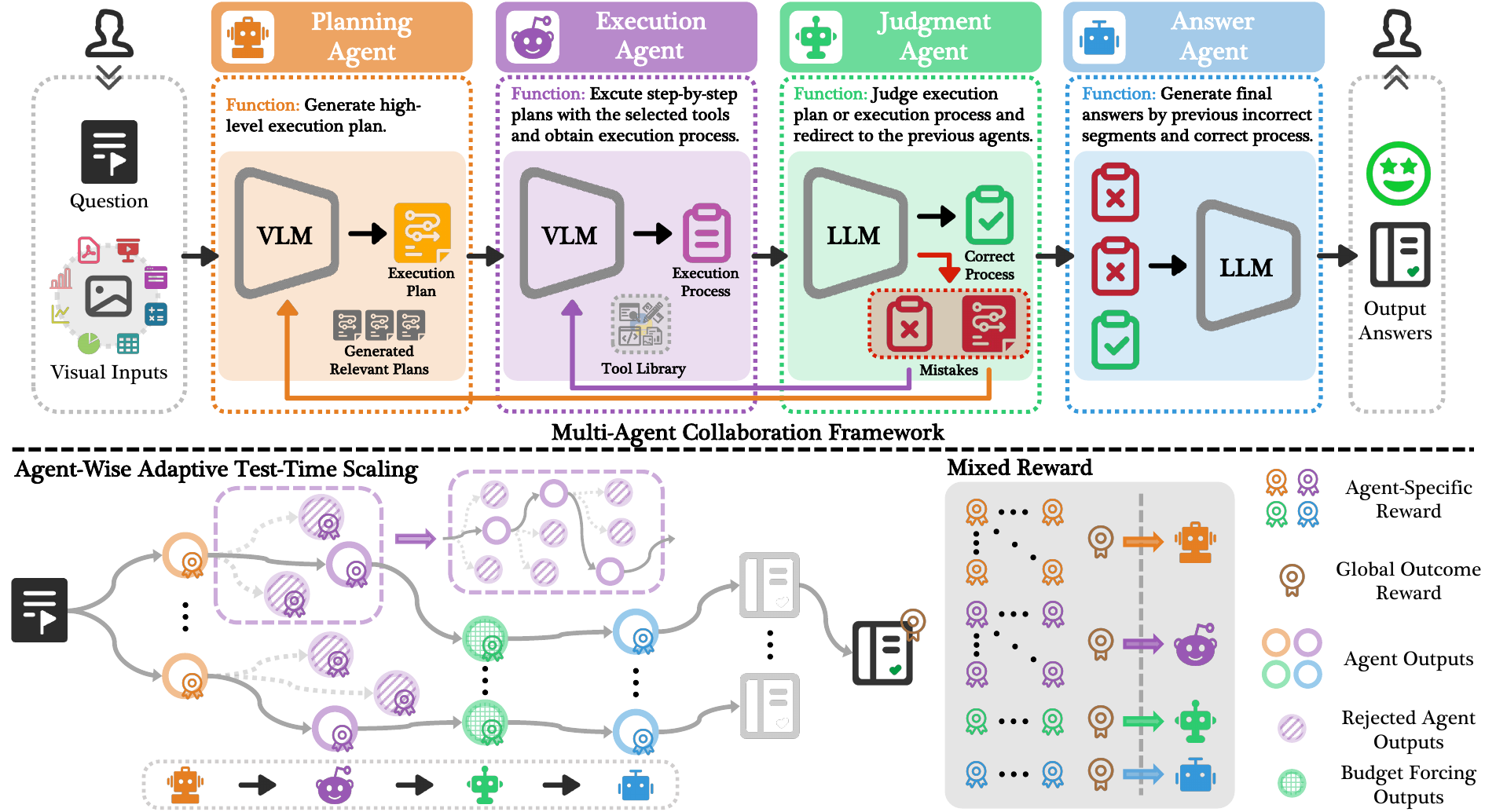}
    \caption{The overview of MACT. The upper part demonstrates our procedural framework, with four tailored and collaborative agents to conduct the process of document analysis. When the judgment agent detects mistakes, it redirects to previous agents for corrections. The lower part illustrates the agent-wise adaptive test-time scaling and mixed reward modeling for the multi-agent framework.}
    \label{overview}
    \vspace{-10pt}
\end{figure*}

\subsection{Multi-Agent Collaboration Framework}
\label{multi_agent_framework}
Document understanding and reasoning constitutes not merely a unilateral perceptual activity, but an expected multistage process \cite{kutsyuruba2023document}: decomposing the question, formulating execution strategies, locating information pertinent to the document, conducting validation with necessary corrections, and synthesizing a final response. 
As shown in the upper part of Fig.~\ref{overview}, MACT consists of four collaborative agents: sequentially planning agent ($\mathcal{A}_{plan}$), execution agent ($\mathcal{A}_{exe}$), judgment agent ($\mathcal{A}_{judg}$), and answer agent ($\mathcal{A}_{ans}$). Each agent is role-specialized, which accomplishes its functionality and then passes to the subsequent agent or outputs the final answers.

\noindent\raisebox{-.3\baselineskip}{\includegraphics[height=1.2\baselineskip]{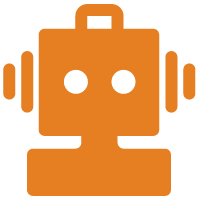}}~\textbf{Planning Agent.} Given a question $Q$ and corresponding visual document inputs $\mathcal{D}$, $\mathcal{A}_{plan}$ mainly focuses on the analysis and decomposition of the original question, making high-level executive plans on the document-based tasks. Before generating the plans for the original task, we generate $\mathit{N_p}$ similar and relevant instances, and their plans simultaneously, which is adapted from analogical prompting principles~\cite{michihiro2024large}. For each generated example and corresponding plan, $\mathcal{A}_{plan}$ generates a distinct execution plan respectively, which could be formulated as:
\begin{align}
        \mathcal{P}_{rel} &= \mathcal{A}_{\text{plan}}\left(\mathcal{Q},\mathcal{D}\right), \\
        \mathcal{P} &= \mathcal{A}_{\text{plan}}\left(\mathcal{Q},\mathcal{P}_{rel},\mathcal{D},\mathcal{M}\right),
\end{align}
where generated relevant plans $\mathcal{P}_{rel} = \left\{r_1,r_2,\dots,r_{\mathit{N_p}}\right\}$, and execution plans $\mathcal{P} = \left\{p_1,p_2,\dots,p_{\mathit{N_p}}\right\}$, which provide different pathways to accomplishment. Besides, $\mathcal{M}$ is the mistake from $\mathcal{A}_{judg}$ for correction, and it is initialized to empty. For each relevant plan and execution plan $p$, it composed of steps $\left\{s_1, s_2, \dots, s_n\right\}$. It should be noted that we only yield high-level execution plans with limited details in this agent. To be more precise, the high-level plan describes the expected targets and requirements of each step, but does not output specific execution processes directly. It ensures that $\mathcal{A}_{plan}$ can better project the overall plan instead of execution details and would not interfere with the execution of $\mathcal{A}_{exe}$. Besides, it avoids premature commitment to specific tools, enabling $\mathcal{A}_{exe}$ to dynamically select optimal resources for diverse tasks.

\vspace{0.3\baselineskip}
\noindent\raisebox{-.3\baselineskip}{\includegraphics[height=1.2\baselineskip]{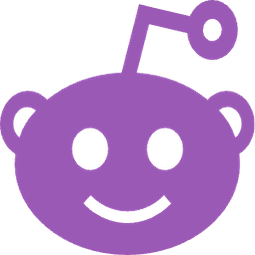}}~\textbf{Execution Agent.} $\mathcal{A}_{exe}$ is designed to execute plans step by step and output the execution process using the selected tools from the tool library $\mathcal{T}=\{t_1,t_2,\dots,t_n\}$ (see in Appendix~{\color{cvprblue}{7.2}}). Specifically, it breaks down the execution plan and each step will be regarded as an execution unit, with information for each unit populated into a template and comprised of: (1) a specific definition; (2) expected target and output; (3) existing inputs or results from the previous step. $\mathcal{A}_{exe}$ then executes the units sequentially, deriving the intended outputs from each. The execution of single step could be presented as:
\begin{equation}
e_i = \mathcal{A}_{\text{exe}}\!\left(\mathcal{Q},\mathcal{D},s_i,\mathcal{T},\mathcal{M}\right).
\end{equation}
Once all steps are completed, it concatenates the full output and the execution processes of all steps in sequence, passing $\mathcal{E} = \left\{ e_1, e_2, \dots, e_n \right\}$ to subsequent agents.



\begin{figure}[t]
    \centering
    \includegraphics[width=0.72\linewidth]{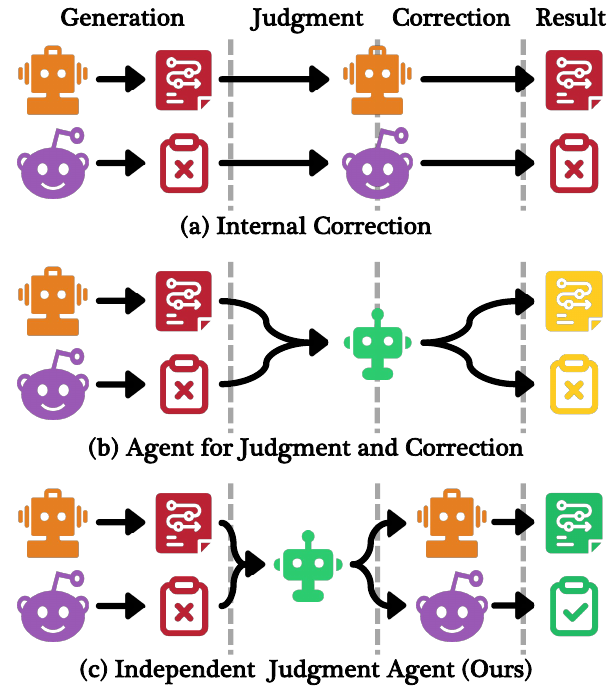}
    \caption{Comparisons of (a) internal correction, (b) an extra agent for both judgment and correction, and (c) our strategy utilizing an independent judgment agent.}
    \label{judg_com}
    \vspace{-1\baselineskip}
\end{figure}

\vspace{0.3\baselineskip}
\noindent\raisebox{-.3\baselineskip}{\includegraphics[height=1.2\baselineskip]{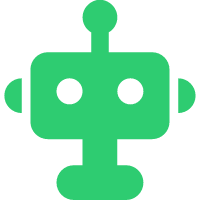}}~\textbf{Judgment Agent.} Document-based visual tasks exhibit high sensitivity to process-related errors~\cite{fitzgerald2012documents}, especially in monolithic models that omit verification and correction procedures. Trivial procedural discrepancies inevitably distort the overall reasoning trajectory, leading to a snowball effect that results in wholly erroneous outputs.
Thus, some multi-agent systems incorporate mechanisms that either (a) internally correct errors within the same agent~\cite{yang2024matplotagent,sun2024pearl} or (b) deploy an additional agent to handle both judgment and correction~\cite{islam2024mapcoder,li2025metal}. However, the (a) internal correction mechanism  struggles to identify most mistakes and may fall into cognitive blind spots, as generation and correction rely on the same model. In contrast, the latter approach (b) requires the agent to possess strong capabilities for both judgment and correction, thereby necessitating models with larger parameters and more complex reward modeling designs. Furthermore, using a separate agent to regenerate mistaken components may result in incompatibilities or even conflicts with existing parts, given the autonomy of the two agents. In RL, the optimization objective for both (a) and (b), as illustrated in Fig.~\ref{judg_com}, is to pass verification. This can lead to strategic production of vague statements or the omission of details, resulting in corrections that appear correct superficially but are actually misleading. And utilizing another agent to regenerate the mistaken part might be incompatible or even causing conflicts with the existing parts because of the independence of two autonomous agents.

To address the limitations of both approaches, we design a simple but effective judgment strategy for multi-agent systems, which introduces an additional judgment agent to separate judgment from correction, thereby fostering specialization of labor. More specifically, $\mathcal{A}_{judg}$ is primarily responsible for assessing the correctness of steps in previously generated execution plans and processes, without engaging in direct correction. The judgment process is summarized as:
\begin{equation}
\mathcal{J}=\mathcal{A}_{judg}\left(\mathcal{Q},\mathcal{P},\mathcal{E}\right),
\end{equation}
where $\mathcal{J}=\left\{flag_{plan},flag_{exe},\mathcal{M}\right\}$, $flag_{plan}=flag_{exe}\in\left\{true,false\right\}$, and $\mathcal{M}$ is the mistake description. If mistakes are detected in any step, $\mathcal{A}_{judg}$ identifies the specific problematic step, provides a brief description of the mistake, and routes the issue to the appropriate agent, either $\mathcal{A}_{plan}$ or $\mathcal{A}_{exe}$, for correction.

For each plan-process pair that is already correct or has been corrected, $\mathcal{A}_{judg}$ forwards them to $\mathcal{A}_{ans}$. This strategy, by decoupling judgment from correction, introduces a neutral judge with reduced subjective bias, whose focus remains solely on judgment rather than correction. Additionally, the reward design for the judgment agent can be intuitively simplified, without evaluating correction outcomes. To prevent infinite correction loops, the maximum number of corrections $N_{c}$ is set to 3.

\vspace{0.3\baselineskip}
\noindent\raisebox{-.3\baselineskip}{\includegraphics[height=1.2\baselineskip]{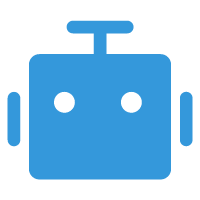}}~\textbf{Answer Agent.} The primary function of $\mathcal{A}_{ans}$ is to output the final answer through both the incorrect processes and the correct process. Counterintuitively, $\mathcal{A}_{ans}$ incorporates both the correct execution process and incorrect segments from prior processes. This facilitates direct focus on modifications within the corrected process, thereby preventing the omission of error-prone details. Additionally, the mistake-correction pair presents a natural and complete logical closed-loop structure, which benefits answer generation based on the execution process. Thus, the final output answer is:
\begin{equation}
\mathcal{O}=\mathcal{A}_{ans}\left(\mathcal{Q},\mathcal{E},\mathcal{M}\right).
\end{equation}

%


\noindent\textbf{Collaboration.} Throughout the full implementation of the multi-agent collaboration framework, visual inputs and questions are first fed into $\mathcal{A}_{plan}$, whose generated plan is then executed by $\mathcal{A}_{exe}$. $\mathcal{A}_{judg}$ judges the correctness the resulting execution plan-process pairs and outputs mistake flags. For correct pairs, $\mathcal{A}_{ans}$ outputs the final answers, while the information of mistakes will be passed to the previous agent $\mathcal{A}_{plan}$ or $\mathcal{A}_{exe}$ for correction, and the procedure repeats. For clarity in illustrating the workflow, we set the number of generated relevant plans to 1 and simplify the test-time scaling designs in Algorithm~\ref{procedure_algorithm}, and these details are elaborated in the subsequent section.


\begin{algorithm}[t]
\caption{Multi-Agent Collaboration Procedure}
\label{procedure_algorithm}
\begin{algorithmic}[1]
\REQUIRE question $\mathcal{Q}$, visual document inputs $\mathcal{D}=\{v_1,v_2,\dots,v_n\}$, planning agent $\mathcal{A}_{plan}$, execution agent $\mathcal{A}_{exe}$, the tool library of the former agent $\mathcal{T}=\{t_1,t_2,\dots,t_n\}$, judgment agent $\mathcal{A}_{judg}$, answer agent $\mathcal{A}_{ans}$, maximum number of corrections $N_{c}$
\ENSURE answer output to the question $\mathcal{O}$

\STATE Initialize the four agents: $\mathcal{A}_{plan}$, $\mathcal{A}_{exe}$, $\mathcal{A}_{judg}$, $\mathcal{A}_{ans}$
\STATE Initialize $N_{c}=3,t=0$
\STATE Initialize prompts formatter $\mathcal{PF}$
\STATE Initialize $flag_{plan}=flag_{exe}=false,\mathcal{M}=empty$

\WHILE{true}
\STATE $p_1\leftarrow\mathcal{PF\left(\mathcal{Q}\right)}$
\STATE $\mathcal{P}_{rel}\leftarrow\mathcal{A}_{plan}\left(\mathcal{D},p_1\right)$ \hfill $\rhd$ Relevant plans
\STATE $p_2\leftarrow\mathcal{PF}\left(\mathcal{Q},\mathcal{P}_{rel},\mathcal{M}\right)$
\STATE $\mathcal{P\leftarrow}\mathcal{A}_{plan}\left(\mathcal{D},p_2\right)$ \hfill $\rhd$ Execution plan
\FOR{each step $s_i$ \textbf{in} $\mathcal{P}$}
\STATE $p_3\leftarrow\mathcal{PF}\left(\mathcal{Q},s_i,\mathcal{T},\mathcal{M}\right)$
\STATE $e_i\leftarrow\mathcal{A}_{exe}\left(\mathcal{D},p_3,\mathcal{T}\right)$ \hfill $\rhd$ Step-wise execution
\ENDFOR
\STATE $\mathcal{E}\leftarrow\left\{e_1,e_2,\dots,e_n\right\}$ \hfill $\rhd$ Execution process
\STATE $p_4\leftarrow\mathcal{PF\left(\mathcal{Q},\mathcal{P},\mathcal{E}\right)}$
\STATE $\mathcal{J}\leftarrow\mathcal{A}_{judg}\left(p_4\right)$ \hfill $\rhd$ Judgment
\STATE $\left\{flag_{plan},flag_{exe},\mathcal{M}\right\}\leftarrow\mathcal{J}$ \hfill $\rhd$ Mistakes
\IF{not $flag_{plan}$ \textbf{and} not $flag_{exe}$ \textbf{or} $t\ge N_{c}$}
    \STATE \textbf{break}
\ELSIF{$flag_{plan}$} 
    \STATE $t \leftarrow t + 1$
    \STATE \textbf{goto} line 8 \hfill $\rhd$ Mistakes in execution plan
\ELSIF{$flag_{exe}$} 
    \STATE $t \leftarrow t + 1$
    \STATE \textbf{goto} line 10 \hfill $\rhd$ Mistakes in execution process
\ENDIF
\ENDWHILE
\STATE $p_5\leftarrow\mathcal{PF\left(\mathcal{Q},\mathcal{E},\mathcal{M}\right)}$
\STATE $\mathcal{O}\leftarrow\mathcal{A}_{ans}\left(p_5\right)$ \hfill $\rhd$ Answer
\end{algorithmic}
\end{algorithm}

\subsection{Agent-Wise Adaptive Test-Time Scaling}
\label{test_time_scaling}

For visual document understanding and reasoning, the key challenge is not visual documents themselves, but the procedural reasoning that links global question decomposition, plan formation, information extraction, and mistake judgment across multiple steps. Monolithic scaling enlarges model capacity uniformly, yet fails to accommodate this structured reasoning process. 
We therefore propose an agent-wise adaptive scaling strategy during inference, dynamically allocating additional computing resources to different functional entities and enabling on-demand allocation during the test time. Existing test-time scaling methods fall into four main categories: parallel scaling, sequential scaling, hybrid scaling, and internal scaling. However, these strategies are originally designed for single models or agents, overlooking the different division of labor among agents and achieving suboptimal performance when applied to multi-agent systems. Accordingly, we propose an agent-wise adaptive scaling strategy tailored to multi-agent architectures, which significantly enhances both agent-specific capabilities and collaborative performance.

For the first three agents with various functions, we customize different test-time scaling strategies to their unique characteristics: (1) \raisebox{-.2\baselineskip}{\includegraphics[height=1.\baselineskip]{figure/plan.png}} Planning agent: In our design of $\mathcal{A}_{plan}$, it naturally provides relevant sample plans as references for formulating multiple execution plans for the original task. We therefore prompt the generation of $N_{p}$ relevant plans independently, yielding $N_{p}$ parallel paths per question. This establishes a basis for scaling subsequent agents within document processing workflows, increasing the probability that at least one reasoning path will produce an accurate answer aligned with document semantics. (2) \raisebox{-.2\baselineskip}{\includegraphics[height=1.\baselineskip]{figure/exe.png}} Execution agent: Given that the execution process is divided by step in $\mathcal{A}_{exe}$, we treat each step as an evaluation node. For each node, the agent produces $\mathit{N_e}$ candidate executions, which are scored using a pretrained reward model. The top-scoring candidate is selected as the base node for subsequent steps, with all others rejected. (3) \raisebox{-.2\baselineskip}{\includegraphics[height=1.\baselineskip]{figure/judg.png}} Judgment agent: To judge the correctness of the execution plans and processes, $\mathcal{A}_{judg}$ requires logical analysis and reasoning to comprehensively and accurately detect mistakes to avoid the error snowballing in document contexts. We therefore adopt the budget forcing scaling method~\cite{muennighoff2025s1} for this agent, which enforces a minimum number of thinking tokens. When token usage falls below the budget, the agent is encouraged to generate additional thinking tokens, thereby promoting accurate judgments. For the answer agent, whose core function lies in synthesizing prior information and generating the final response, test-time scaling confers merely marginal improvements.

\subsection{Mixed Reward Modeling}
\label{mixed_reward}
According to the latest research~\cite{setlur2025scaling,zhang2025survey}, adding RL strategy, \textit{e.g.}, reward modeling, to optimize test-time scaling is significantly superior to fine-tuning or training-free methodologies. To optimize our proposed procedural scaling by elevating task-agent-centric and collaborative performance, we design a mixed reward strategy that synthesizes agent-specific rewards and global outcome-driven reward signals. 
Given that the four agents in the framework have distinct functions and reward signal preferences, we first incorporate agent-tailored rewards:
\begin{align}
    r_{plan/exe} &= \mathcal{R}_{prm}\left({s}_{i}/{e}_i\mid\mathcal{Q},\mathcal{D},\mathcal{P}/\mathcal{E}\right), \\
    r_{judg/ans} &= \mathcal{R}_{orm}\left(\mathcal{J}/\mathcal{O}\mid\mathcal{Q}\right),
\end{align}
where are $\mathcal{R}_{prm}$ and $\mathcal{R}_{orm}$ are multi-modal process reward and outcome reward models, respectively. The former yields step-wise process rewards, providing instant feedback for each step and hierarchical rewards for $\mathcal{A}_{plan}$ and $\mathcal{A}_{exe}$, and the latter generates single reward signal for each output of $\mathcal{A}_{judg}$ and $\mathcal{A}_{ans}$. 
Furthermore, we apply a global reward based on the final selected path of the four agents:
\begin{equation}
    r_{global} = \mathcal{R}_{orm}\left(\left\{\mathcal{P},\mathcal{E},\mathcal{J},\mathcal{O}\right\}\mid\mathcal{Q},\mathcal{D}\right),
\end{equation}
which reinforces incentives for correct paths and moderately mitigates agent self-interest. Since individual agents tend to optimize their own local rewards, the integration of a global objective effectively alleviates such self-serving tendencies. Please refer to Appendix~{\color{cvprblue}{7.3}} for details.

%


\begin{table*}[t]
\centering
\caption{Comparisons of our MACT and other counterparts on 15 benchmarks. The table cells with \textcolor{green!40}{green}, \textcolor{blue!40}{blue}, and \textcolor{gray!40}{gray} colors denote the best, the second best, and the third best values. Results with - indicate exceeding the maximum context token limit or not being equipped with a specific ability. $\triangle, \Box, \Diamond $ correspond to the base models of three variants of MACT for comparisons.}
\setlength{\tabcolsep}{0.9mm}{
\resizebox{1\linewidth}{!}{
\begin{tabular}{c|l|cccc|cc|cc|cc|cc|ccc|c}
\toprule
   \multicolumn{2}{c|}{} & \multicolumn{10}{c|}{\textbf{Document}} & \multicolumn{5}{c|}{\textbf{Non-Document}} & \multirow{3}{*}{\textbf{Avg.}} \\ 
   \multicolumn{2}{c|}{} & \multicolumn{4}{c|}{\textbf{Text}} & \multicolumn{2}{c|}{\textbf{Webpage}} & \multicolumn{2}{c|}{\textbf{Chart}} & \multicolumn{2}{c|}{\textbf{Table}} & \multicolumn{2}{c|}{\textbf{General}} & \multicolumn{3}{c|}{\textbf{Mathematical}} &  \\ 
  \multicolumn{2}{c|}{}  & DVQA & DUDE & SVQA & MMLong & VisMRC & InfVQA & ChartQA & CharXiv & TabVQA & TabBen & SciQA & RealQA & MVista & MVision & MVerse & \\ \midrule
\multirow{27}{*}{\rotatebox{90}{\textbf{Generalist}}} & \multicolumn{1}{c}{\textbf{Closed-Source}} \\

& GPT-4o-latest  & 93.1 & 52.7 & 81.0 & 40.3 & 86.4 & 79.2 & 86.5 & 78.7 & 64.2 & 51.9 & \cellcolor{green!20}{\textbf{83.3}} & 76.2 & 62.4 & 30.7 & 39.4 & 67.2 \\ 
& Claude-3.7-Sonnet & 94.0 & 58.1 & \cellcolor{gray!20}{\textbf{83.6}} & 33.9 & 82.5 & 75.3 & \cellcolor{green!20}{\textbf{92.2}} & 83.5 & 70.3 & 54.7 & 80.4 & 71.9 & 69.7 & 41.2 & 45.0 & 69.1 \\
& Gemini-2.0-Pro & 91.8 & 54.3 & 78.9 & 32.2 & \cellcolor{gray!20}{\textbf{91.4}} & 81.6 & 88.8 & 83.1 & 71.2 & \cellcolor{gray!20}{\textbf{59.9}} & \cellcolor{gray!20}{\textbf{80.9}} & 70.5 & 74.8 & \cellcolor{blue!20}{\textbf{54.2}} & \cellcolor{blue!20}{\textbf{56.6}} & 71.3 \\ \cline{2-18} \noalign{\vskip 3pt}
& \multicolumn{1}{c}{\textbf{Size $<$ 20B}} & \\
& DeepSeek-VL2-4.5B  & 78.8 & - & - & - & 70.4 & 65.3 & 74.0 & 46.9 & 30.8 & - & 66.6 & 60.7 & 29.5 & 6.9 & 7.4 & - \\
& LLaVA-OneVision-7B-si & 86.6 & 48.4 & 52.9 & 5.3 & 76.1 & 74.8 & 82.3 & 54.3 & 47.9 & 33.3 & 72.8 & 62.0 & 57.9 & 17.9 & 19.1 & 52.8 \\
& Qwen2.5-VL-7B-Instruct$^\triangle$ & 94.6 & 62.3 & 76.2 & 22.4 & 86.8 & 82.8 & 87.5 & 68.2 & 60.4 & 49.2 & 74.0 & 68.4 & 67.8 & 25.5 & 40.8 & 64.5 \\
& MiMo-VL-7B-SFT$^\Box$ & 92.9 & 62.3 & 74.2 & 27.5 & 87.3 & 82.7 & 86.1 & 72.0 & 57.5 & 52.5 & 75.1 & 67.9 & 70.7 & \cellcolor{gray!20}{\textbf{46.9}} & 53.3 & 67.3 \\
& InternVL3-8B-Instruct$^\Diamond$ & 91.1 & 60.1 & 72.9 & 28.6 & 84.4 & 78.5 & 85.4 & 67.1 & 59.7 & 52.4 & 76.4 & 66.7 & 65.5 & 28.8 & 38.9 & 63.8  \\ 
& Llama-3.2-11B-Vision-Instruct & 87.3 & 45.0 & 66.5 & 8.8 & 79.6 & 73.8 & 79.6 & 63.6 & 50.4 & 44.6 & 76.9 & 60.7 & 56.5 & 16.3 & 32.3 & 56.1 \\
& LlaVa-1.6-vicuna-13B & 79.0 & 42.9 & 65.6 & 7.9 & 77.9 & 74.7 & 76.6 & 52.5 & 24.9 & 39.2 & 70.7 & 58.1 & 50.2 & 11.6 & 27.5 & 50.6\\
& InternVL3-14B-Instruct & 90.7 & 61.0 & 74.5 & 28.9 & 87.1 & 83.1 & 86.2 & 75.1 & 58.2 & 51.4 & 77.1 & 66.4 & 66.4 & 36.6 & 43.5 & 65.7 \\
& Ovis2-16B & 92.4 & 59.6 & 73.8 & 29.0 & 88.5 & 78.1 & 87.9 & 70.7 & 59.5 & 46.8 & 72.5 & 67.9 & 69.6 & 35.6 & 45.0 & 65.1 \\ \cline{2-18} \noalign{\vskip 3pt}
& \multicolumn{1}{c}{\textbf{30B $\le$ Size $<$ 40B}} & \\
& Qwen2.5-VL-32B-Instruct  & 94.9 & 66.4 & 75.0 & 22.9 & 86.4 & 83.8 & 87.7 & 75.9 & 62.7 & 51.0 & 74.8 & 71.7 & 65.8 & 32.6 & 46.4 & 66.5 \\
& LlaVa-1.6-34B & 87.8 & 50.1 & 69.5 & 9.9 & 80.1 & 78.4 & 85.3 & 60.4 & 32.9 & 44.4 & 76.2 & 67.0 & 56.7 & 15.8 & 17.0 & 55.4\\ 
& Ovis2-34B  & 93.6 & 63.1 & 72.4 & 31.0 & 85.6 & 79.1 & 88.9 & 71.9 & 61.6 & 47.1 & 72.0 & 70.7 & 74.0 & 31.4 & 49.5 & 66.1 \\
& InternVL3-38B-Instruct & 95.0 & 61.7 & 76.3 & 30.2 & 89.9 & 84.6 & 88.0 & 79.8 & 63.1 & 54.1 & 77.5 & 72.8 & 75.2 & 34.0 & 47.6 & 68.7 \\ \cline{2-18} \noalign{\vskip 3pt}
& \multicolumn{1}{c}{\textbf{70B $\le$ Size $<$ 100B}} & \\
& LLaVA-OneVision-72B-si & 91.7 & 62.7 & 63.9 & 10.6 & 81.4 & 75.2 & 86.1 & 63.8 & 52.8 & 41.5 & 75.1 & 70.5 & 72.5 & 25.2 & 27.0 & 60.0 \\
& Qwen2.5-VL-72B-Instruct  & \cellcolor{gray!20}{\textbf{95.7}} & 67.0 & 82.4 & 24.9 & 90.3 & 87.5 & 89.5 & 80.6 & 68.4 & 58.1 & 74.4 & 74.8 & 77.6 & 39.8 & 46.2 & 70.5 \\
& InternVL3-78B-Instruct & 95.3 & 68.9 & 80.7 & 32.4 & 90.7 & 86.2 & 89.6 & 77.9 & 65.5 & 57.9 & 78.0 & \cellcolor{gray!20}{\textbf{77.9}} & 79.3 & 44.3 & 48.7 & 71.6 \\
& Llama-3.2-90B-Vision-Instruct & 91.4 & 58.6 & 76.7 & 19.5 & 86.0 & 82.3 & 86.3 & 74.4 & 60.1 & 49.4 & 78.5 & 68.8 & 58.7 & 23.4 & 35.0 & 63.3 \\ \midrule 
 
\multirow{7}{*}{\rotatebox{90}{\textbf{Specialist}}} & MMCA-7B & 68.9 & - & - & - & 59.0 & 54.9 & 64.4 & 47.5 & - & - & 56.6 & 52.6 & - & - & - & -\\
& UReader-7B & 72.3 & - & - & - & 70.7 & 66.6 & 67.0 & 52.0 & - & - & 60.4 & 55.7 & - & - & - & - \\
& mPLUG-DocOwl2-8B & 87.8 & 46.5 & - & - & 76.1 & 72.2 & 80.8 & 52.4 & 25.7 & 19.8  & 65.7 & 60.0 & 41.3 & 9.2 & 14.4 & - \\
& TextMonkey-10B  & 77.2 & 36.8 & - & - & 74.5 & 68.1 & 72.5 & 52.0 & 21.7 & 16.9 & 65.9 & 63.4 & 33.7 & 5.4 & 11.7 & -\\
& M3DocRAG-10B & 79.2 & 39.5 & 57.9 & 29.2 & 71.8 & 67.3 & 73.7 & 56.5 & 22.5 & 16.5 & 65.4 & 58.9 & 41.8 & 14.8 & 20.8 & 47.7 \\
& CogAgent-17B & 80.9 & - & - & - & 70.3 & 65.3 & 69.9 & 54.3 & 21.7 & - & 58.8 & 52.6 & - & - & - & - \\
& MDocAgent-39B & 86.4 & 58.4 & 68.6 & 36.7 & 82.8 & 74.4 & 82.8 & 70.2 & 58.8 & 48.7 & 69.8 & 64.8 & 46.6 & 20.1 & 29.3 & 59.9 \\ \midrule 
\multirow{6}{*}{\rotatebox{90}{\textbf{Ours}}} & \textbf{MACT}  & \cellcolor{green!20}{\textbf{96.6}} & \cellcolor{green!20}{\textbf{72.5}} & \cellcolor{green!20}{\textbf{85.3}} & \cellcolor{gray!20}{\textbf{43.7}} & \cellcolor{blue!20}{\textbf{92.0}} & \cellcolor{green!20}{\textbf{89.4}} & \cellcolor{blue!20}{\textbf{91.9}} & \cellcolor{blue!20}{\textbf{85.2}} & \cellcolor{blue!20}{\textbf{74.0}} & 57.2 & 78.5 & \cellcolor{blue!20}{\textbf{77.7}} & \cellcolor{gray!20}{\textbf{81.2}} & 41.8 & \cellcolor{gray!20}{\textbf{54.7}} & \cellcolor{gray!20}{\textbf{74.8}} \\
& \multicolumn{1}{l|}{\textbf{-Qwen2.5-VL-Series-24B}$^\triangle$} & \multicolumn{1}{r}{\cellcolor{white}{+2.0}} & \multicolumn{1}{r}{\cellcolor{white}{+10.2}} & \multicolumn{1}{r}{\cellcolor{white}{+9.1}} & \multicolumn{1}{r|}{\cellcolor{white}{+21.3}} & \multicolumn{1}{r}{\cellcolor{white}{+5.2}} & \multicolumn{1}{r|}{\cellcolor{white}{+6.6}} & \multicolumn{1}{r}{\cellcolor{white}{+4.4}} & \multicolumn{1}{r|}{\cellcolor{white}{+17.0}} & \multicolumn{1}{r}{\cellcolor{white}{+13.6}} & \multicolumn{1}{r|}{\cellcolor{white}{+8.0}} & \multicolumn{1}{r}{\cellcolor{white}{+4.5}} & \multicolumn{1}{r|}{\cellcolor{white}{+9.3}} & \multicolumn{1}{r}{\cellcolor{white}{+13.4}} & \multicolumn{1}{r}{\cellcolor{white}{+16.3}} & \multicolumn{1}{r|}{\cellcolor{white}{+13.9}} & \multicolumn{1}{r}{\cellcolor{white}{+10.3}} \\
& \textbf{MACT} & 94.4 & \cellcolor{blue!20}{\textbf{70.8}} & \cellcolor{blue!20}{\textbf{83.9}} & \cellcolor{green!20}{\textbf{47.4}} & \cellcolor{green!20}{\textbf{93.8}} & \cellcolor{blue!20}{\textbf{88.6}} & \cellcolor{gray!20}{\textbf{91.4}} & \cellcolor{green!20}{\textbf{87.2}} & \cellcolor{gray!20}{\textbf{71.6}} & \cellcolor{green!20}{\textbf{62.7}} & 79.2 & 76.1 & \cellcolor{green!20}{\textbf{85.4}} & \cellcolor{green!20}{\textbf{60.1}} & \cellcolor{green!20}{\textbf{65.3}} & \cellcolor{green!20}{\textbf{77.2}} \\ 
& \multicolumn{1}{l|}{\textbf{-MiMo-VL-Series-28B}$^\Box$} & \multicolumn{1}{r}{\cellcolor{white}{+1.5}} & \multicolumn{1}{r}{\cellcolor{white}{+8.5}} & \multicolumn{1}{r}{\cellcolor{white}{+9.7}} & \multicolumn{1}{r|}{\cellcolor{white}{+19.9}} & \multicolumn{1}{r}{\cellcolor{white}{+6.5}} & \multicolumn{1}{r|}{\cellcolor{white}{+5.9}} & \multicolumn{1}{r}{\cellcolor{white}{+5.3}} & \multicolumn{1}{r|}{\cellcolor{white}{+15.2}} & \multicolumn{1}{r}{\cellcolor{white}{+14.1}} & \multicolumn{1}{r|}{\cellcolor{white}{+10.2}} & \multicolumn{1}{r}{\cellcolor{white}{+4.1}} & \multicolumn{1}{r|}{\cellcolor{white}{+8.2}} & \multicolumn{1}{r}{\cellcolor{white}{+14.7}} & \multicolumn{1}{r}{\cellcolor{white}{+13.2}} & \multicolumn{1}{r|}{\cellcolor{white}{+12.0}} & \multicolumn{1}{r}{\cellcolor{white}{+9.9}} \\
& \textbf{MACT} & \cellcolor{blue!20}{\textbf{96.1}} & \cellcolor{gray!20}{\textbf{69.8}} & 81.3 & \cellcolor{blue!20}{\textbf{46.5}} & 91.0 & \cellcolor{gray!20}{\textbf{87.7}} & 90.6 & \cellcolor{gray!20}{\textbf{85.0}} & \cellcolor{green!20}{\textbf{75.3}} & \cellcolor{blue!20}{\textbf{61.0}} & \cellcolor{blue!20}{\textbf{81.3}} & \cellcolor{green!20}{\textbf{80.1}} & \cellcolor{blue!20}{\textbf{83.8}} & 45.8 & 54.0 & \cellcolor{blue!20}{\textbf{75.3}} \\
& \multicolumn{1}{l|}{\textbf{-InternVL3-Series-28B}$^\Diamond$} & \multicolumn{1}{r}{\cellcolor{white}{+5.0}} & \multicolumn{1}{r}{\cellcolor{white}{+9.7}} & \multicolumn{1}{r}{\cellcolor{white}{+8.4}} & \multicolumn{1}{r|}{\cellcolor{white}{+17.9}} & \multicolumn{1}{r}{\cellcolor{white}{+6.6}} & \multicolumn{1}{r|}{\cellcolor{white}{+9.2}} & \multicolumn{1}{r}{\cellcolor{white}{+5.2}} & \multicolumn{1}{r|}{\cellcolor{white}{+17.9}} & \multicolumn{1}{r}{\cellcolor{white}{+15.6}} & \multicolumn{1}{r|}{\cellcolor{white}{+8.6}} & \multicolumn{1}{r}{\cellcolor{white}{+4.9}} & \multicolumn{1}{r|}{\cellcolor{white}{+13.4}} & \multicolumn{1}{r}{\cellcolor{white}{+18.3}} & \multicolumn{1}{r}{\cellcolor{white}{+17.0}} & \multicolumn{1}{r|}{\cellcolor{white}{+15.1}} & \multicolumn{1}{r}{\cellcolor{white}{+11.5}} \\
\bottomrule
\end{tabular}}}
\label{comparison}
\vspace{-1\baselineskip}
\end{table*}

\section{Experiments}
More details and descriptions about the training pipelines, datasets, evaluations and prompts for each agent are provided in Appendix~{\color{cvprblue}{8.1}}, {\color{cvprblue}{8.2}}, {\color{cvprblue}{8.3}}, and {\color{cvprblue}{8.4}}, respectively.

\subsection{Training Pipeline}
Since the functions and optimization goals of each agent are independent and diverse, we design a two-stage SFT and RL pipeline for MACT. To understand visual inputs and better accomplish their labors, we choose VLMs for the $\mathcal{A}_{plan}$ and $\mathcal{A}_{exe}$, while LLMs for the $\mathcal{A}_{judg}$ and $\mathcal{A}_{ans}$. For all the four agents, we initiate with pretrained models, and we select three groups of small-parameter base models for different variants of MACT: \textbf{(1) Qwen2.5-VL series based}~\cite{bai2025qwen2,qwen2025qwen25technicalreport}: Qwen2.5-VL-7B-Instruct and Qwen2.5-7B/3B-Instruct; \textbf{(2) MiMo-VL series based}~\cite{coreteam2025mimovl,xia2025mimo}: MiMo-VL-7B-SFT and MiMo-7B-SFT;  \textbf{(3) InternVL3 series based}~\cite{zhu2025internvl3}: InternVL3-9B/8B/2B-Instruct. In the first SFT stage, we initially train a tuned 11B/7B/7B VLM on the selected document-based or non-document-based datasets, mixing data with or without CoT. It aims to enhance their visual understanding and reasoning abilities, providing more robust models for future multi-agent collaboration. Next, we fine-tune an 8B/7B/7B LLM on judgment labels generated via GPT-4o~\cite{hurst2024gpt} and rule-based verifications. Finally, we fine-tune another 3B/3B/7B LLM on the outputs from preceding agents and ground-truths to enhance its summary ability. The fine-tuned VLM are applied as $\mathcal{A}_{plan}$ and $\mathcal{A}_{exe}$ respectively in the subsequent stage, while the two LLMs function as the $\mathcal{A}_{judg}$ and $\mathcal{A}_{ans}$. In the second RL stage, we generate reward signals based on pretrained reward models and optimize our model via GRPO~\cite{shao2024deepseekmath}. Process reward model VisualPRM~\cite{wang2025visualprm} is utilized for the step-by-step reward signals of $\mathcal{A}_{plan}$ and $\mathcal{A}_{exe}$, while Skywork-VL-Reward~\cite{wang2025skywork} is used for the rewards generation of $\mathcal{A}_{judg}$ and $\mathcal{A}_{ans}$.

\subsection{Benchmarks} 
To comprehensively and objectively evaluate our model's document-based capabilities, we selected 15 datasets encompassing four real-world document categories and two non-document general categories. The four document types are comprised of: \textbf{(1) Text-based:} DocVQA~\cite{mathew2021docvqa}, DUDE~\cite{van2023document}, SlideVQA~\cite{tanaka2023slidevqa}, MMLongBench-Doc~\cite{ma2025mmlongbench}; \textbf{(2) Webpage-based:} VisualMRC~\cite{tanaka2021visualmrc}, InfographicVQA~\cite{mathew2022infographicvqa}; \textbf{(3) Chart-based:} ChartQA~\cite{masry2022chartqa}, CharXiv~\cite{wang2024charxiv}; \textbf{(4) Table-based:} TableVQA-Bench~\cite{kim2024tablevqa}, and TableBench~\cite{wu2025tablebench}. To validate our document-centric paradigm without sacrificing capabilities in general domains, two non-document types are involved: \textbf{(1) General:} ScienceQA~\cite{saikh2022scienceqa}, RealWorldQA~\cite{2024XAIrealworld}; \textbf{(2) Mathematical:} MathVista~\cite{lu2024mathvista}, Math-Vision~\cite{wang2024measuring}, and MathVerse~\cite{zhang2024mathverse}. We adhere to the original training and testing splits and use the testing splits as evaluation benchmarks. Each instance in these datasets consists of a visual input sequence and a question, spanning difficulty levels from easy to hard and encompassing various question types for comprehensive evaluation.

\subsection{Evaluations}
For other reproducible models, we utilize LMMs-Eval~\cite{zhang2024lmms} for fair comparisons on both natively supported and our registered benchmarks. For most benchmarks, we employ GPT-4o~\cite{hurst2024gpt} as a judge model to evaluate the correctness of the generated answers based on LMMs-Eval. For the remaining benchmarks, we utilize their original evaluation metrics, \textit{e.g.}, ANLS and F1, as detailed in Tab.~{\color{cvprblue}{8}}.


\section{Results and Discussions}

\subsection{Main Results}
We evaluate MACT across 15 benchmarks: 10 document-based benchmarks to assess its document understanding performance; 2 general benchmarks and 3 mathematical benchmarks are designed to verify that it retains full capabilities in non-document-centric and reasoning task settings. For comparison, we select state-of-the-art (SOTA) methods, encompassing both generalist and specialist models, and categorize them by parameter size.

\vspace{0.3\baselineskip}
\noindent \textbf{Benchmark Results.} As shown in Tab.~\ref{comparison}, the MACT-MiMo-VL-Series-28B variant delivers the best average performance, followed by MACT-InternVL3-Series-28B and MACT-Qwen2.5-VL-Series-24B, which rank second and third, respectively. Despite having fewer than 30B parameters, MACT models outperform all comparative methods with under 100B parameters, as well as closed-source models. Notably, MACT-MiMo-VL-Series-28B achieves significant improvements of 5.6\% and 5.9\% over the top-performing open-source and closed-source models in terms of the average scores. Additionally, the three variants top 13 of the 15 benchmarks, with MACT-MiMo-VL-Series-28B leading on seven. Notably, in MMLongBench-Doc, which features the longest visual context, and across the three mathematical reasoning benchmarks, MACT-MiMo-VL-Series-28B outperforms the second-highest scorer by 7.1\%, 10.6\%, 5.9\%, and 8.7\%, respectively.
These results highlight that document-based tasks, far from being monolithic constructs, are better suited to procedural scaling than to monolithic models with larger parameters, supporting flexible adaptation to functional entities and computational resource allocation.

\vspace{0.3\baselineskip}
\noindent \textbf{Comparisons with Base Models.} As shown in Fig.~\ref{comparison_to_base}, radar charts compare three variants of our proposed MACT framework with their base models and larger models from the same series. It is observed that MACT models significantly outperform their base models by 10.3\%, 9.9\%, and 11.5\% on average across the 15 benchmarks. Besides, compared to their corresponding larger monolithic models, \textit{i.e.}, Qwen2.5-VL-72B-Instruct and InternVL3-78B-Instruct, our MACT-Qwen2.5-VL-Series-24B and MACT-InternVL3-Series-28B achieve average performance gains of at least 6.6\% and 3.7\%, respectively, with more pronounced improvements observed on long-context and reasoning tasks. These results demonstrate that our method, equipped with procedural scaling paradigm, is superior to monolithic paradigm in the document-based tasks and also more general domains.

\subsection{Additional Quantitative Analysis}
To further demonstrate the effectiveness of our proposed procedural scaling paradigm, we perform ablation experiments and analyses of scaling components. Experiments here are conducted using MACT-Qwen2.5-VL-Series-24B, and the training pipelines and training data are consistent with the main experiment. We select three lowest-performing benchmarks, distinguished by long visual contexts and reasoning-intensive demands, and the average performance across all benchmarks, aiming to quantify their impacts on both specialized capabilities.

\begin{table}[t]
\centering
\caption{Results of ablations on multi-agent collaboration, agent-wise adaptive test-time scaling, and mixed reward modeling.}
\setlength{\tabcolsep}{0.9mm}{
\resizebox{0.99\linewidth}{!}{
\begin{tabular}{l|cccc}
\toprule
 & \textbf{MMLong} & \textbf{TabBen} & \textbf{MVision} & \textbf{Avg.} \\ \midrule
Monolithic & 32.5  & 50.8 & 32.4 & 66.2 \\
w/o Multi-Agent Collaboration & 24.7 & 44.4 & 26.0 & 58.6 \\
w/o Agent-Wise Adaptive Scaling & 34.9 & 50.8 & 34.5  & 71.1  \\ 
w/o Mixed Reward Modeling & 38.3 & 54.2 & 36.7 & 71.4  \\\midrule
\textbf{MACT} & \textbf{43.7} & \textbf{57.2} & \textbf{41.8} & \textbf{74.8} \\ \bottomrule
\end{tabular}}}
\label{ablation}
\end{table}

\begin{table}[t]
\centering
\caption{Results of different combinations of agents.}
\setlength{\tabcolsep}{0.9mm}{
\resizebox{0.92\linewidth}{!}{
\begin{tabular}{ccc|cccc}
 \toprule
 $\mathcal{A}_{plan}$ + $\mathcal{A}_{exe}$ & $\mathcal{A}_{judg}$ & $\mathcal{A}_{ans}$ & \textbf{MMLong} & \textbf{TabBen} & \textbf{MVision} & \textbf{Avg.} \\ \midrule
 \CheckmarkBold & & & 36.3 & 48.5 & 34.9  & 68.4 \\
 \CheckmarkBold & \CheckmarkBold & & 42.2 & 56.7 & 40.2 & 73.9 \\
 \CheckmarkBold & & \CheckmarkBold & 36.6 & 48.6 & 35.4  & 68.8 \\ \midrule
 \CheckmarkBold & \CheckmarkBold & \CheckmarkBold & \textbf{43.7} & \textbf{57.2} & \textbf{41.8} & \textbf{74.8} \\ \bottomrule
\end{tabular}}}
\label{agent_ablation}
\vspace{-1\baselineskip}
\end{table}

\vspace{0.3\baselineskip}
\noindent\textbf{Ablation Studies.} The ablation settings are as follows: (1) Monolithic: Use a monolithic model to directly execute tasks and output answers, with particular scaling and reward modeling retained. (2) w/o multi-agent collaboration: Use a single agent, which incorporates all prompts from our four proposed agents as a unified workflow, with our proposed scaling and reward strategies retained. (3) w/o agent-wise adaptive test-time scaling: Retain the multi-agent collaborative framework and apply the mixed reward strategy directly, without test-time scaling. (4) w/o mixed reward modeling: Retain the multi-agent collaborative framework and agent-wise adaptive test-time scaling, without reward signals. As reported in Tab.~\ref{ablation}, our document-centric procedural scaling paradigm, outperforming the monolithic system by 8.6\%. 
Notably, the integration of all functionalities into a single model results in suboptimal performance, even outperforming the base model negatively, which indicates the inherent limitation of monolithic paradigms in document understanding and reasoning. Both the agent-wise adaptive test-time scaling strategy and mixed reward modeling contribute to an average accuracy increase of at least 3\%. Specifically, the former yields more remarkable gains on reasoning tasks in documents than average ones, with particularly striking improvements that underscore how adaptive scaling reduces cognitive overload.

\begin{table}[t]
\centering
\caption{Results of different setting of test-time scaling strategies. }
\setlength{\tabcolsep}{0.9mm}{
\resizebox{0.85\linewidth}{!}{
\begin{tabular}{l|cccc}
\toprule
& \textbf{MMLong} & \textbf{TabBen} & \textbf{MVision} & \textbf{Avg.} \\ \midrule
No Scaling & 34.9 & 50.8 & 34.5 & 71.1 \\
Parallel Scaling & 39.1 & 55.2 & 37.5 & 72.0 \\
Sequential Scaling & 41.3 & 54.7 & 40.0 & 72.4 \\
Hybrid Scaling & 39.8 & 55.5 & 39.4 & 73.0 \\ 
Internal Scaling & 41.8 & 55.9 & 41.6 & 72.3 \\ \midrule
\textbf{Agent-Wise Adaptive} & \textbf{43.7} & \textbf{57.2} & \textbf{41.8} & \textbf{74.8} \\  \bottomrule
\end{tabular}}}
\label{tts_analysis}
\end{table}

\begin{table}[t]
\centering
\caption{Results of different setting of reward modeling. }
\setlength{\tabcolsep}{0.9mm}{
\resizebox{0.85\linewidth}{!}{
\begin{tabular}{l|cccc}
\toprule
& \textbf{MMLong} & \textbf{TabBen} & \textbf{MVision} & \textbf{Avg.} \\ \midrule
No Reward & 38.3 & 54.2 & 36.7 & 71.4 \\
Agent-Specific Reward & 42.8 & 56.3 & 40.5 & 72.7  \\
Global Reward & 34.1 & 51.6 & 35.1 & 70.2\\ \midrule
\textbf{Mixed Reward} & \textbf{43.7} & \textbf{57.2} & \textbf{41.8} & \textbf{74.8} \\  \bottomrule
\end{tabular}}}
\vspace{-1\baselineskip}
\label{reward_analysis}
\end{table}

\begin{figure}[t]
    \centering 
    \begin{subfigure}[b]{0.23\textwidth} 
        \centering
        \includegraphics[width=\textwidth]{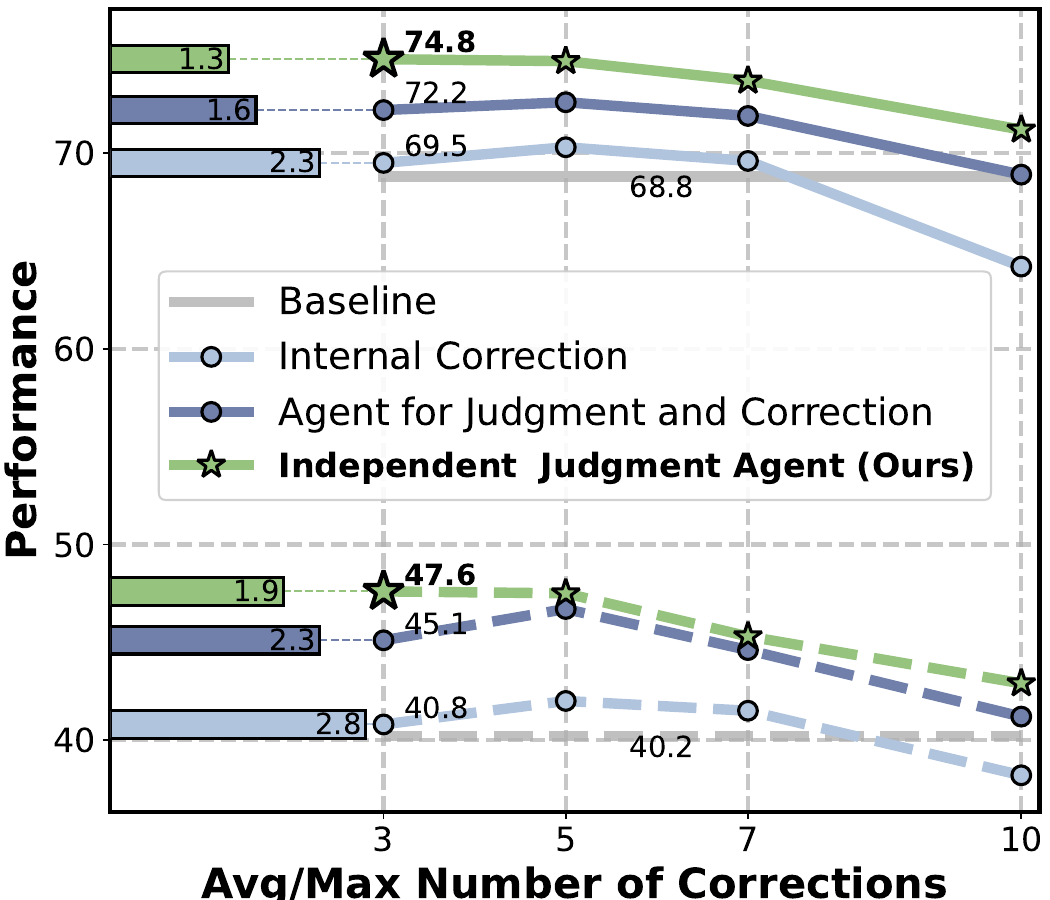}
        \caption{Analysis of Corrections}
        \label{correct_com}       
    \end{subfigure} 
    \hfill 
    \begin{subfigure}[b]{0.23\textwidth}
        \centering
        \includegraphics[width=\textwidth]{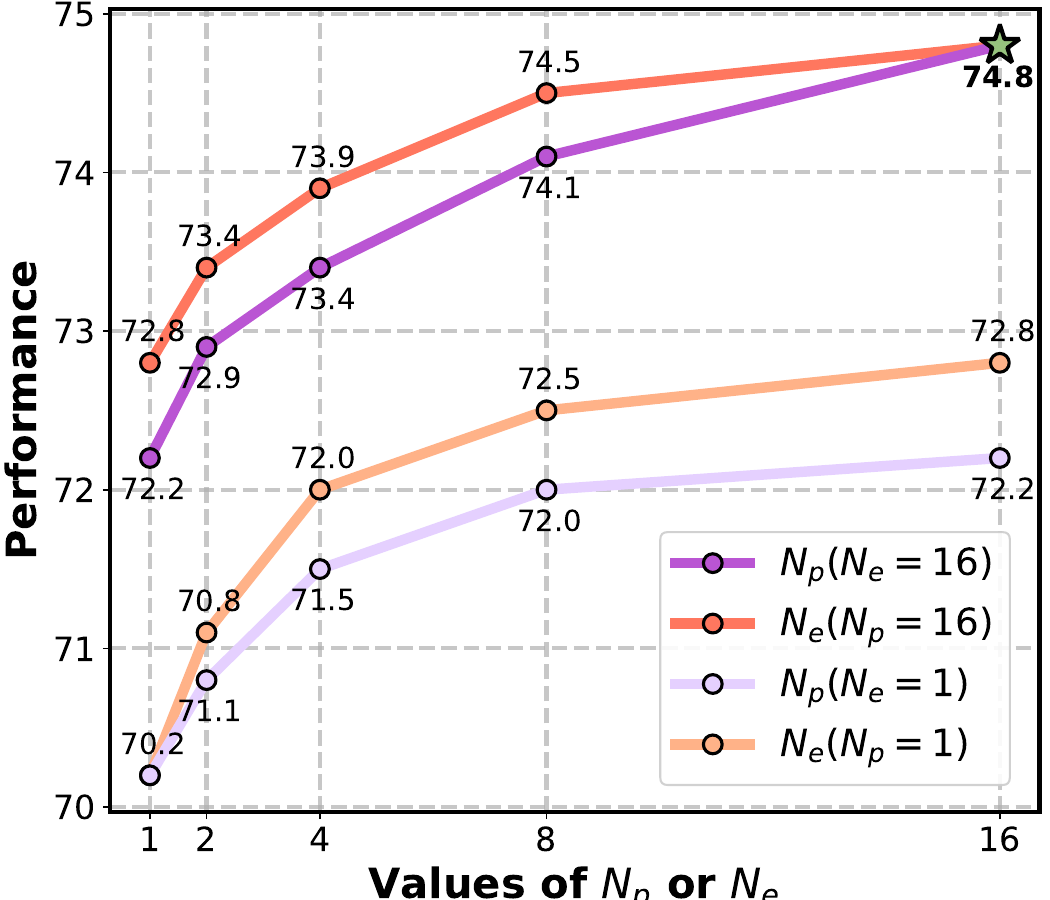}
        \caption{Impact of $\mathit{N_p}$ and $\mathit{N_e}$}
        \label{scaling_com}
    \end{subfigure}

    \caption{(a) Line graph shows the impact of various maximum numbers of corrections, with solid and dashed lines denoting average values across all and three selected benchmarks, respectively. The bar charts show the average judgment numbers when the maximum is set to 3. (b) The line graphs represent the impacts of the number of generated plans $\mathit{N_p}$ and candidate executions $\mathit{N_e}$.} 
    \label{ablation_pdf}  
    \vspace{-1\baselineskip}
\end{figure}

\vspace{0.3\baselineskip}
\noindent\textbf{Analysis of Multi-Agent Collaboration.} To assess the utility of each agent and the effectiveness their procedural collaboration, we perform additional experiments on different agent combinations, which are demonstrated in Tab.~\ref{agent_ablation}. The combination of $\mathcal{A}_{plan}$ and $\mathcal{A}_{exe}$, as a basic procedural multi-agent system, results in a 3.9\% average increase over the base model, while adding $\mathcal{A}_{judg}$ further boosts performance by 5.5\%. $\mathcal{A}_{ans}$ additionally improves the scores by 0.9\% on the average. Additionally, as in Fig.~\ref{correct_com}, we compare our judgment agent strategy with the other two counterparts, \textit{i.e.}, internal correction and agent for both judgment and correction. Our proposed strategy outperforms the others by at least 2.6\% on average, while requiring an average of 0.3 fewer corrections. Moreover, our method achieves optimal correction results when the maximum number of corrections $N_{c}$ is set to 3, whereas the alternatives require $N_{c}$ to be set to 5, highlighting its greater accuracy and efficiency. It could also be observed that when setting a higher $N_{c}$, the performance will not improve continuously, primarily because excessive corrections may confuse the agent and obscure correct answers.

\vspace{0.3\baselineskip}
\noindent\textbf{Analysis of Agent-Wise Adaptive Test-Time Scaling and Mixed Reward modeling.} We perform experiments on our proposed agent-wise adaptive test-time scaling and mixed reward modeling. For the former, we benchmark our method with the other four effective test-time scaling strategies, as listed in Tab.~\ref{tts_analysis}. Here, we select budget forcing~\cite{muennighoff2025s1} as the internal scaling.
Our agent-wise adaptive scaling strategy based on the computational resources and cognitive load of each agent, outperforms all four existing strategies and improves the average scores by at least 1.8\%. As shown in Fig.~\ref{scaling_com}, higher values of $\mathit{N_p}$ and $\mathit{N_e}$ increase the likelihood of finding the correct answers, which leads to more attempts at plans and execution steps. For the latter, we separately use the agent-wise reward and global reward for comparison, as reported in Tab.~\ref{reward_analysis}. Although the improvement from global reward is relatively limited, it avoids the selfishness of agents that only use agent-specific reward, yielding a further 1.3\% increase in average scores.

\section{Conclusion}
Our proposed MACT is an innovative multi-agent collaboration framework with adaptive test-time scaling strategy for visual document understanding and reasoning, shifting the monolithic scaling into procedural scaling. It comprises four collaborative agents that deconstructs the workflow of document analysis, achieving effective role division and collaboration. Notably, the judgment agent strategy outperforms existing self-correction mechanisms while requiring fewer corrections. Additionally, agent-wise adaptive test-time scaling and mixed reward modeling further reduce the cognitive overload adaptively. Extensive comparative experiments and supplementary analysis validate the superiority of our MACT. This procedural paradigm constitutes a meaningful exploration of multi-agent frameworks and adaptive test-time scaling strategies for document-based scenarios and tasks.
{
    \small
    \bibliographystyle{ieeenat_fullname}
    \bibliography{main}
}


\end{document}